\documentclass{article}

 \usepackage[dblblindworkshop, final]{neurips_2025}

\usepackage[utf8]{inputenc} % allow utf-8 input
\usepackage[T1]{fontenc}    % use 8-bit T1 fonts
\usepackage{hyperref}       % hyperlinks
\usepackage{url}            % simple URL typesetting
\usepackage{booktabs}       % professional-quality tables
\usepackage{amsfonts}       % blackboard math symbols
\usepackage{nicefrac}       % compact symbols for 1/2, etc.
\usepackage{microtype}      % microtypography
\usepackage{xcolor}         % colors
\usepackage{listings}
\usepackage{graphicx}
\usepackage{titlesec}

\title{Fuzzy, Symbolic, and Contextual: Enhancing LLM Instruction via Cognitive Scaffolding}
\workshoptitle{CogInterp: Interpreting Cognition in Deep Learning Models}

\author{%
  Vanessa Figueiredo\thanks{https://www.vanesfi.com/} \\
  Department of Computer Science\\
  University of Regina\\
  Regina, SK \\
  \texttt{vanessa.figueiredo@uregina.ca} \\
}

\begin{document}

\maketitle

\begin{abstract}
We study how prompt-level inductive biases influence the cognitive behavior of large language models (LLMs) in instructional dialogue. We introduce a symbolic scaffolding method paired with a short-term memory schema designed to promote adaptive, structured reasoning in Socratic tutoring. Using controlled ablation across five system variants, we evaluate model outputs via expert-designed rubrics covering scaffolding, responsiveness, symbolic reasoning, and conversational memory. We present preliminary results using an LLM-based evaluation framework aligned to a cognitively grounded rubric. This enables scalable, systematic comparisons across architectural variants in early-stage experimentation. The preliminary results show that our full system consistently outperforms baseline variants. Analysis reveals that removing memory or symbolic structure degrades key cognitive behaviors, including abstraction, adaptive probing, and conceptual continuity. These findings support a processing-level account in which prompt-level cognitive scaffolds can reliably shape emergent instructional strategies in LLMs.
\end{abstract}

\section{Introduction}
LLMs excel in linguistic fluency but struggle with dynamic reasoning \citep{liu2023lost, gao-etal-2021-making} and maintaining task-relevant state \citep{li2025language, pink2024assessing} over multiple turns, especially when user needs evolve or ambiguity arises. Inspired by cognitive theories of control and scaffolding, we propose a modular fuzzy, symbolic framework that supports interpretable, adaptive behavior through prompt-level symbolic reasoning. Symbolic representations can offer a tractable interface between LLMs and structured reasoning spaces \citep{patel2022mapping}, grounding learning processes in interpretable scaffolds.

Building on prior work \citep{figueiredo2025fuzzy} that introduced a two-layer scaffolded prompting structure grounded in Vygotskian theory \citep{vygotsky_mind_1978}, we augment this framework with a third layer: a structured short-term memory schema (\texttt{short\_term\_schema.json}). This schema tracks session variables (e.g., learner profile, task type, scaffolding strategy) across turns, enabling runtime modulation of strategies without fine-tuning.

Rather than scaling models or relying on retrieval, our method embeds cognitive scaffolds that act as prompt-level inductive bias. To evaluate this system, we focus on Socratic-style tutoring tasks, which simulate one-on-one instructional dialogues where the assistant must guide learners through probing questions, interpret confusion, and adapt its strategy across turns. This setting offers a cognitively rich testbed for evaluating behavioral coherence and symbolic control in LLMs. While human ratings are planned for future work, we argue that structured LLM scoring can serve as a high-throughput behavioral screening method for early experimental stages. Results suggest that the fuzzy, symbolic framework improves coherence, responsiveness, and scaffolding adaptivity compared to ablated baselines.

\textbf{Our contributions:}
\begin{itemize}
  \item A modular natural language boundary framework operationalizing fuzzy, symbolic scaffolding.
  \item A short-term memory schema enabling turn-by-turn cognitive control.
  \item A prompt-level symbolic loop for real-time strategy modulation.
  \item A rubric-based evaluation framework adapted from instructional science.
\end{itemize}

\section{Related Work}
\textbf{Schemas as Controllers.} Schema theory \citep{alma993888043803476, rumelhart:blocks} offers a cognitive lens on structure-based generalization. Unlike classical systems (e.g., ACT-R, SOAR) \citep{AndersonJohnR.1998TACo, LairdJohn2012TSca}, we treat schemas both as memory and runtime behavior controllers.

\textbf{Memory-Augmented LMs.} Efforts to improve long-range coherence include retrieval-augmented methods (e.g., kNN-LMs \citep{khandelwal2019generalization}, RAG \citep{lewis2020retrieval}). However, these approaches prioritize access over structure, yielding unorganized and task-agnostic recall. Other studies \citep{chan2022data} reveal that in-context learning reflects dataset distributional artifacts more than principled reasoning. We address this by introducing a symbolic short-term schema that encodes context persistently and interprets it adaptively across dialogue turns.

\textbf{Fuzzy Reasoning.} Human reasoning is inherently graded and context-sensitive. Fuzzy logic \citep{zadeh-1994-soft} models this uncertainty and has seen successful use in fuzzy trees \citep{ishibuchi2002effect}, hesitant sets \citep{torra2010hesitant}, and linguistic evaluations \citep{Herrera-Viedma-2021-revisiting}. Most prompting pipelines, however, reduce ambiguity to binary outcomes. Our fuzzy scaffolding strategy enables graded feedback and dynamic modulation of instructional support, closely mirroring human-like pedagogical reasoning.

\textbf{Interpretability.} Post-hoc methods \citep{lipton2017mythosmodelinterpretability} often fail to reflect causal reasoning. We instead enable symbolic interpretability via schema-guided reasoning, modulating behavior through explicit, updatable controllers. Our approach aligns with compositional studies in synthetic agents \citep{okawa2023compositional}, extending them toward symbolic reasoning in LLMs.

\textbf{Toward Operational Cognition in LLMs}
Prior work treats cognition as emergent or post-hoc. We propose an interface model: a symbolic control loop combining boundary prompts, fuzzy heuristic rules schema, and structured memory. This supports interpretability, session-level coherence, and dynamic adaptivity, hallmarks of cognitive control. Rather than probing black-box behavior, we design for it, embedding structure at inference time to produce interpretable, pedagogically grounded responses.

\section{Problem Definition}

\begin{figure}[h]
  \centering
  \includegraphics[width=0.6\linewidth]{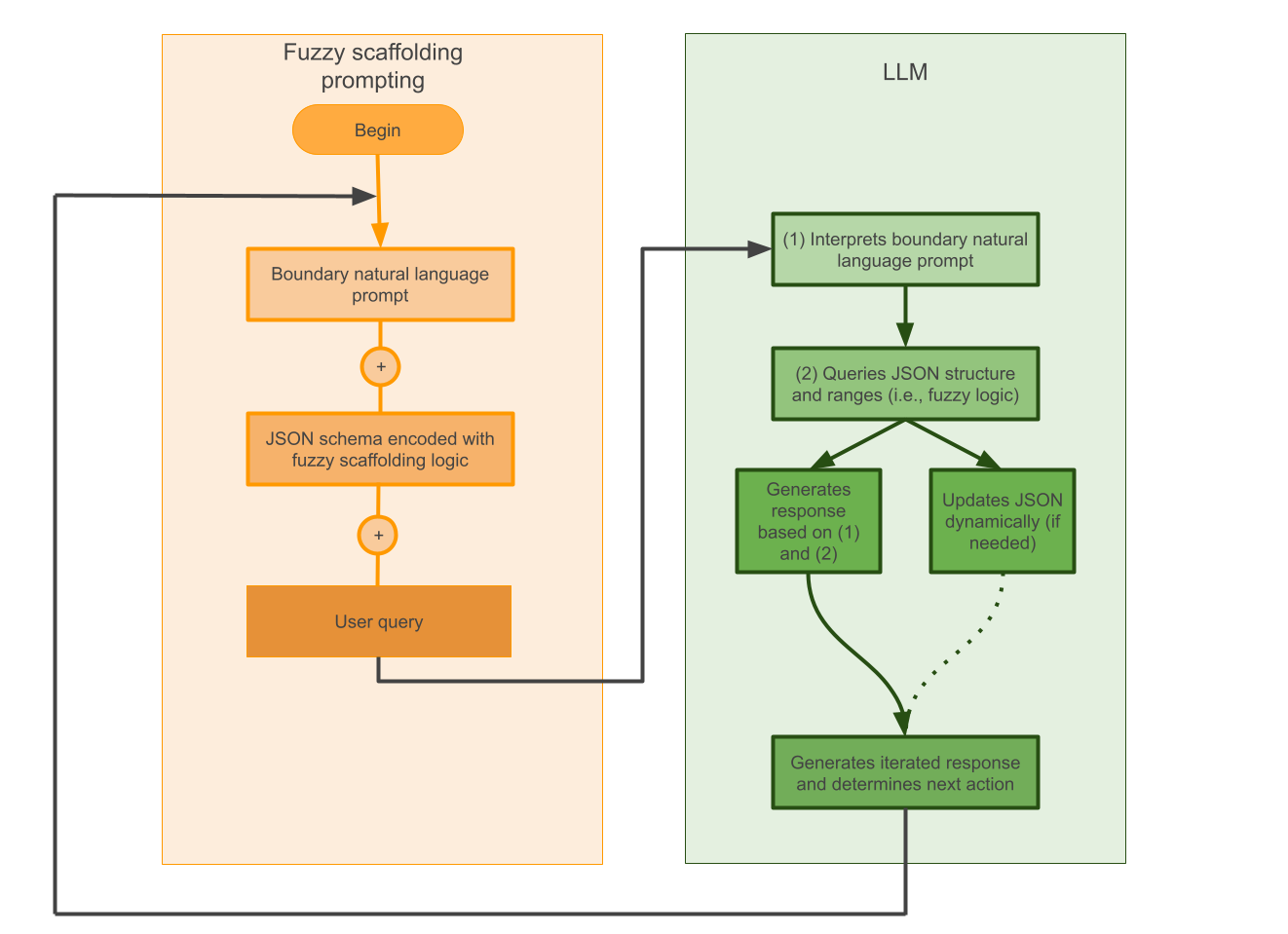}
  \caption{Runtime prompting loop with symbolic scaffolding and memory. Each turn integrates boundary prompts, fuzzy heuristics, and symbolic memory updates to produce adaptive, context-aware behavior.}
  \label{fig:inference-diagram}
\end{figure}

LLMs often struggle to maintain context, adapt reasoning strategies, and handle uncertainty across multi-turn interactions. We propose a framework that introduces \textit{prompt-level inductive biases}: structured, inference-time constraints that guide model behavior without modifying its underlying architecture. Particularly, we ask: how can an LLM (1) \textbf{represent} and update symbolic control states based on user behavior; (2) \textbf{adapt} its reasoning strategies (e.g., tone, scaffolding, vocabulary) to reflect learner understanding; and (3) \textbf{preserve} interpretability and coherence across dialogue turns (Figure~\ref{fig:inference-diagram}).

\subsection*{Symbolic Scaffolding Framework}

\textbf{Symbolic scaffolding} is a structured prompting method that guides model reasoning in a manner similar to a human tutor offering step-by-step support \citep{vygotsky_mind_1978}. The framework consists of three interlinked layers: (1) a \textbf{boundary prompt}, which defines the model’s role, domain, and instructional goals to frame the interaction context; (2) a \textbf{fuzzy schema}, which encodes heuristic, graded interpretations of learner state, enabling soft modulation of instructional strategies under uncertainty (e.g., distinguishing between “partially understood” and “completely unclear” states); and (3) a \textbf{symbolic memory schema}, which tracks key variables such as prior misconceptions, strategy history, or knowledge progression, allowing the model to behave adaptively across turns without retraining.

While the fuzzy schema does not implement classical fuzzy logic systems (e.g., it lacks formal membership functions or defuzzification), it draws from fuzzy reasoning principles to support human-interpretable, flexible responses.

\subsection*{Inference-Time Control Loop}

At each turn, the model follows a structured inference loop: it (1) parses the user’s input to infer knowledge state or confusion; (2) consults the fuzzy schema to determine an appropriate scaffolding strategy; (3) generates a context-aware response based on boundary constraints and heuristic logic; and (4) updates the symbolic memory with relevant state changes (e.g., misconceptions corrected or learning progress observed), which inform subsequent interactions.

\subsection*{Illustrative Example}

Consider a learner who says: \textit{“I think the moon changes shape because clouds move in front of it.”} The \textbf{boundary prompt} ensures the model responds as a science tutor with a supportive, pedagogically grounded tone. The \textbf{fuzzy schema} recognizes this as a partial misconception and triggers a mid-level scaffolding move, acknowledging the learner's observation while prompting correction. The \textbf{symbolic memory schema} records the misunderstanding (“clouds cause phases”) for follow-up in future turns.

This structured inference-time control enables interpretable, adaptive behavior without architectural changes. Full implementation details are provided in Appendix~\ref{appendix:framework}.

\section{Experimental Setup}
We assess our fuzzy, symbolic framework within the domain of Socratic tutoring. We selected two domains: \textit{global warming} (grade 7 readability level) and \textit{moon phases} (grade 11 readability level). Each scenario was seeded with a simulated learner utterance reflecting epistemic uncertainty (e.g., "I'm not sure I understand…"), followed by 5-7 turns of user-assistant interaction.

We curated 10 distinct dialogue scenarios (5 per domain). Each prompt condition generated a complete multi-turn dialogue for each scenario, producing a total of \textbf{255 assistant responses} across 50 dialogues (5 conditions × 10 dialogues). All tutoring responses were generated using the \textbf{LLaMA 3.1–8B model} served locally via \texttt{Ollama}, ensuring reproducibility and architectural consistency.

\begin{table}[h]
\centering
\caption{Dialogue Turns by Condition and Topic}
\label{tab:turn-counts}
\begin{tabular}{lcc}
\toprule
\textbf{Condition} & \textbf{Global Warming} & \textbf{Moon Phases} \\
\midrule
C0 (Full)         & 27 & 26 \\
C1 (No-Memory)    & 25 & 26 \\
C2 (No-Fuzzy)     & 26 & 27 \\
C3 (No-Boundary)  & 25 & 26 \\
C4 (Vanilla)      & 26 & 27 \\
\bottomrule
\end{tabular}
\end{table}

\subsection{Prompting Conditions}

To isolate the functional contributions of each symbolic control layer, we define five experimental prompting variants:

\begin{itemize}
    \item \textbf{C0 (Full):} Full symbolic prompting (boundary prompt + fuzzy schema + short-term memory).
    \item \textbf{C1 (No-Memory):} Ablates memory updates—no persistent control state across turns.
    \item \textbf{C2 (No-Fuzzy):} Removes fuzzy logic scaffolding—uses static heuristics for support.
    \item \textbf{C3 (No-Boundary):} Omits task-role framing and adaptation rules.
    \item \textbf{C4 (Vanilla):} Baseline instruction-following prompt with no symbolic control.
\end{itemize}

While each dialogue scenario was seeded with identical initial learner utterances, the number of turns per dialogue varied slightly across prompt conditions. This variability reflects the dynamic nature of Socratic interactions and is an emergent consequence of the model’s scaffolding behavior.

\subsection{Evaluation Protocol}
We adopted a rubric-based evaluation framework to assess cognitive adaptivity. Each assistant utterance was scored along five dimensions: (1) \textbf{Scaffolding quality}: clarity and appropriateness of pedagogical support; (2) \textbf{Contextual responsiveness}: relevance to learner input and turn history; (3) \textbf{Helpfulness}: contribution toward conceptual understanding; (4) \textbf{Symbolic strategy use}: evidence of structured reasoning or planning; (5) \textbf{Memory of conversation}: continuity and adaptation across turns.

All ratings were produced via GPT-4 using a structured evaluation rubric. Scores were normalized to a 1–5 scale, and each judgment included a free-text rationale for interpretability. The rubric draws from established frameworks in instructional design and cognitive modeling \citep{vanlehn2011relative, wood1976role, koedinger2012knowledge} and are available in Appendix \ref{appendix:rubric}.

\section{Results}
We computed mean ratings per condition and ran repeated-measures ANOVAs across prompt variants for each dimension. Where omnibus tests revealed significant differences, we performed Tukey HSD post-hoc tests to identify pairwise contrasts. All significance values and effect sizes were extracted to assess which symbolic components contributed to improved adaptivity and coherence (see Appendix \ref{appendix:results} for detailed results). A one-way ANOVA across the five conditions (C0–C4) revealed significant effects of prompting framework on all rated dimensions ($p < 0.01$), with the \textbf{C0 full} achieving the highest mean scores across the board. Post-hoc Tukey HSD tests confirmed that C0 significantly outperformed vanilla prompting (C4) in every metric ($p < 0.001$), and also yielded statistically significant improvements over partial ablations (C1–C3) in dimensions tied to their respective modules.

\begin{itemize}
    \item \textbf{Memory module (C1 ablation):} Removing memory led to significant declines in multi-turn coherence and adaptivity ($p = 0.004$).
    \item \textbf{Fuzzy schema (C2 ablation):} Excluding fuzzy logic reduced scaffolding quality and symbolic reasoning ($p = 0.012$).
    \item \textbf{Boundary prompt (C3 ablation):} Omitting explicit role/task framing resulted in diminished contextual responsiveness and interpretability ($p = 0.002$).
\end{itemize}

Scores were consistent across both tutoring domains, suggesting that the symbolic scaffolding framework generalizes across content. In particular, \textbf{symbolic strategy use} and \textbf{memory-aware modulation} showed low variance between \textit{moon phases} and \textit{global warming} topics, supporting the domain-agnostic nature of the framework.

LLM-generated justifications revealed interpretable distinctions between conditions. For C0 (Full), the assistant frequently referenced learner progress, offered specific scaffolding techniques (e.g., visual aids, simplification), and reflected on earlier conversation turns. In contrast, vanilla prompts (C4) often produced disjointed or overly general replies. Partial ablations exhibited mixed strategies, failing to sustain a coherent control loop.

While these results are based on preliminary LLM-based evaluations, the rubric was explicitly designed to support future integration of human expert ratings. This synthetic evaluation phase played a critical role in refining our framework, allowing us to stabilize the core architecture and validate the effectiveness of the fuzzy heuristic rules. A revised version of the system, informed by these insights, is currently under development and will soon undergo human evaluation to enable more nuanced and context-sensitive assessments.

\section{Planned Experiments and Applicability}
While our present work focuses on instructional dialogue, the framework itself is modular and domain-agnostic. Its structure, combining symbolic scaffolding, fuzzy heuristic rules, and memory-based modulation, lends itself naturally to a wide range of applications that involve complex, multi-turn reasoning and non-binary decision spaces. In parallel work, we have successfully adapted a similar prompting framework to support instructional dialogues in other school subjects, including mathematics and social sciences \citep{10.1145/3765284}. These extensions retained the fuzzy scaffolding schema and boundary prompt principles, demonstrating the framework’s flexibility across knowledge domains and pedagogical genres. Additionally, we explored grade-sensitive entertainment dialogues, such as educational puzzles tuned to readability levels using the Flesch-Kincaid metric, illustrating the framework’s capacity to scale across both curricular and informal learning contexts. 
We are currently applying this approach to procedural dialogue generation in games \citep{figueiredo2025symbolically}, where the ability to track narrative state and adapt conversational strategies across turns is essential for maintaining immersion and believability. Beyond gaming, we see strong potential for this architecture in domains such as health communication, where patient interactions often involve uncertainty and require adaptive explanation strategies; scientific explanation assistants, which must handle varied prior knowledge and scaffold abstract reasoning; and customer service systems, where dynamic personalization and context tracking are increasingly expected.

\section{Discussion and Conclusion}
Symbolic prompting offers a transparent and adaptive approach to reasoning in LLMs. Our structured memory schema enables real-time cognitive control without model retraining, encoding interpretability and behavioral policies directly within prompts, unlike retrieval-based methods. While the present study is limited to synthetic users and LLM-based evaluations, it provides a reproducible framework for operationalizing cognitive control. Future work will extend this approach to human evaluations and hybrid symbolic–neural memory systems. Overall, symbolic scaffolding emerges as a practical pathway toward more trustworthy and cognitively grounded language agents.

{
\small
\bibliographystyle{plainnat}
\bibliography{main}

\begin{thebibliography}{25}
\providecommand{\natexlab}[1]{#1}
\providecommand{\url}[1]{\texttt{#1}}
\expandafter\ifx\csname urlstyle\endcsname\relax
  \providecommand{\doi}[1]{doi: #1}\else
  \providecommand{\doi}{doi: \begingroup \urlstyle{rm}\Url}\fi

\bibitem[Anderson and Lebiere(1998)]{AndersonJohnR.1998TACo}
John~R. Anderson and Christian Lebiere.
\newblock \emph{The Atomic Components of Thought}.
\newblock Psychology Press, Mahwah, N.J, 1 edition, 1998.
\newblock ISBN 0805828168.

\bibitem[Bartlett(1995)]{alma993888043803476}
Frederic~C. Bartlett.
\newblock \emph{Remembering : a study in experimental and social psychology.}
\newblock The Cambridge psychological library. University Press, Cambridge, 1995.

\bibitem[Chan et~al.(2022)Chan, Santoro, Lampinen, Wang, Singh, Richemond, McClelland, and Hill]{chan2022data}
Stephanie Chan, Adam Santoro, Andrew Lampinen, Jane Wang, Aaditya Singh, Pierre Richemond, James McClelland, and Felix Hill.
\newblock Data distributional properties drive emergent in-context learning in transformers.
\newblock \emph{Advances in neural information processing systems}, 35:\penalty0 18878--18891, 2022.

\bibitem[Figueiredo(2025{\natexlab{a}})]{10.1145/3765284}
Vanessa Figueiredo.
\newblock Designing smarter conversational agents for kids: Lessons from cognitive work and means-ends analyses.
\newblock \emph{ACM Trans. Comput.-Hum. Interact.}, September 2025{\natexlab{a}}.
\newblock ISSN 1073-0516.
\newblock \doi{10.1145/3765284}.
\newblock URL \url{https://doi.org/10.1145/3765284}.
\newblock Just Accepted.

\bibitem[Figueiredo(2025{\natexlab{b}})]{figueiredo2025fuzzy}
Vanessa Figueiredo.
\newblock A fuzzy logic prompting framework for large language models in adaptive and uncertain tasks.
\newblock \emph{pre-print arXiv:2508.06754}, 2025{\natexlab{b}}.

\bibitem[Figueiredo(2025{\natexlab{c}})]{figueiredo2025symbolically}
Vanessa Figueiredo.
\newblock Symbolically scaffolded play: Balancing control and improvisation in llm-powered npc dialogue.
\newblock \emph{pre-print arXiv}, 2025{\natexlab{c}}.

\bibitem[Gao et~al.(2021)Gao, Fisch, and Chen]{gao-etal-2021-making}
Tianyu Gao, Adam Fisch, and Danqi Chen.
\newblock Making pre-trained language models better few-shot learners.
\newblock In Chengqing Zong, Fei Xia, Wenjie Li, and Roberto Navigli, editors, \emph{Proceedings of the 59th Annual Meeting of the Association for Computational Linguistics and the 11th International Joint Conference on Natural Language Processing (Volume 1: Long Papers}, pages 3816--3830. Association for Computational Linguistics, Aug 2021.
\newblock URL \url{https://aclanthology.org/2021.acl-long.295/}.

\bibitem[Herrera-Viedma et~al.(2021)Herrera-Viedma, Palomares, Li, Cabrerizo, Dong, Chiclana, and Herrera]{Herrera-Viedma-2021-revisiting}
Enrique Herrera-Viedma, Iv{\'a}n Palomares, Cong-Cong Li, Francisco~Javier Cabrerizo, Yucheng Dong, Francisco Chiclana, and Francisco Herrera.
\newblock Revisiting fuzzy and linguistic decision making: Scenarios and challenges for making wiser decisions in a better way.
\newblock \emph{IEEE Transactions on Systems, Man, and Cybernetics: Systems}, 51\penalty0 (1):\penalty0 191--208, 2021.

\bibitem[Ishibuchi and Nakashima(2002)]{ishibuchi2002effect}
Hisao Ishibuchi and Tomoharu Nakashima.
\newblock Effect of rule weights in fuzzy rule-based classification systems.
\newblock \emph{IEEE transactions on fuzzy systems}, 9\penalty0 (4):\penalty0 506--515, 2002.

\bibitem[Khandelwal et~al.(2019)Khandelwal, Levy, Jurafsky, Zettlemoyer, and Lewis]{khandelwal2019generalization}
Urvashi Khandelwal, Omer Levy, Dan Jurafsky, Luke Zettlemoyer, and Mike Lewis.
\newblock Generalization through memorization: Nearest neighbor language models.
\newblock \emph{arXiv preprint arXiv:1911.00172}, 2019.

\bibitem[Koedinger et~al.(2012)Koedinger, Corbett, and Perfetti]{koedinger2012knowledge}
Kenneth~R Koedinger, Albert~T Corbett, and Charles Perfetti.
\newblock The knowledge-learning-instruction framework: Bridging the science-practice chasm to enhance robust student learning.
\newblock \emph{Cognitive science}, 36\penalty0 (5):\penalty0 757--798, 2012.

\bibitem[Laird(2012)]{LairdJohn2012TSca}
John Laird.
\newblock \emph{The Soar cognitive architecture}.
\newblock MIT Press, London, England, 2012.
\newblock ISBN 1280499273.

\bibitem[Lewis et~al.(2020)Lewis, Perez, Piktus, Petroni, Karpukhin, Goyal, K{\"u}ttler, Lewis, Yih, Rockt{\"a}schel, et~al.]{lewis2020retrieval}
Patrick Lewis, Ethan Perez, Aleksandra Piktus, Fabio Petroni, Vladimir Karpukhin, Naman Goyal, Heinrich K{\"u}ttler, Mike Lewis, Wen-tau Yih, Tim Rockt{\"a}schel, et~al.
\newblock Retrieval-augmented generation for knowledge-intensive nlp tasks.
\newblock \emph{Advances in neural information processing systems}, 33:\penalty0 9459--9474, 2020.

\bibitem[Li et~al.(2025)Li, Guo, and Andreas]{li2025language}
Belinda~Z Li, Zifan~Carl Guo, and Jacob Andreas.
\newblock (how) do language models track state?
\newblock \emph{arXiv preprint arXiv:2503.02854}, 2025.

\bibitem[Lipton(2017)]{lipton2017mythosmodelinterpretability}
Zachary~C. Lipton.
\newblock The mythos of model interpretability, 2017.
\newblock URL \url{https://arxiv.org/abs/1606.03490}.

\bibitem[Liu et~al.(2023)Liu, Lin, Hewitt, Paranjape, Bevilacqua, Petroni, and Liang]{liu2023lost}
Nelson~F Liu, Kevin Lin, John Hewitt, Ashwin Paranjape, Michele Bevilacqua, Fabio Petroni, and Percy Liang.
\newblock Lost in the middle: How language models use long contexts.
\newblock \emph{arXiv preprint arXiv:2307.03172}, 2023.

\bibitem[Okawa et~al.(2023)Okawa, Lubana, Dick, and Tanaka]{okawa2023compositional}
Maya Okawa, Ekdeep~S Lubana, Robert Dick, and Hidenori Tanaka.
\newblock Compositional abilities emerge multiplicatively: Exploring diffusion models on a synthetic task.
\newblock \emph{Advances in Neural Information Processing Systems}, 36:\penalty0 50173--50195, 2023.

\bibitem[Patel and Pavlick(2022)]{patel2022mapping}
Roma Patel and Ellie Pavlick.
\newblock Mapping language models to grounded conceptual spaces.
\newblock In \emph{International conference on learning representations}, 2022.

\bibitem[Pink et~al.(2024)Pink, Vo, Wu, Mu, Turek, Hasson, Norman, Michelmann, Huth, and Toneva]{pink2024assessing}
Mathis Pink, Vy~A Vo, Qinyuan Wu, Jianing Mu, Javier~S Turek, Uri Hasson, Kenneth~A Norman, Sebastian Michelmann, Alexander Huth, and Mariya Toneva.
\newblock Assessing episodic memory in llms with sequence order recall tasks.
\newblock \emph{arXiv preprint arXiv:2410.08133}, 2024.

\bibitem[Rumelhart(1980)]{rumelhart:blocks}
David~E. Rumelhart.
\newblock Schemata: {T}he building blocks of cognition.
\newblock In R.~J. Spiro, B.~C. Bruce, and W.~F. Brewer, editors, \emph{Theoretical Issues in Reading Comprehension}, pages 33--58. Erlbaum, Hillsdale, NJ, 1980.

\bibitem[Torra(2010)]{torra2010hesitant}
Vicen{\c{c}} Torra.
\newblock Hesitant fuzzy sets.
\newblock \emph{International journal of intelligent systems}, 25\penalty0 (6):\penalty0 529--539, 2010.

\bibitem[VanLehn(2011)]{vanlehn2011relative}
Kurt VanLehn.
\newblock The relative effectiveness of human tutoring, intelligent tutoring systems, and other tutoring systems.
\newblock \emph{Educational psychologist}, 46\penalty0 (4):\penalty0 197--221, 2011.

\bibitem[Vygotsky and Cole(1978)]{vygotsky_mind_1978}
L.~S. Vygotsky and Michael Cole.
\newblock \emph{Mind in society: the development of higher psychological processes}.
\newblock Harvard University Press, Cambridge, 1978.
\newblock ISBN 978-0-674-57628-5 978-0-674-57629-2.

\bibitem[Wood et~al.(1976)Wood, Bruner, and Ross]{wood1976role}
David Wood, Jerome~S Bruner, and Gail Ross.
\newblock The role of tutoring in problem solving.
\newblock \emph{Journal of child psychology and psychiatry}, 17\penalty0 (2):\penalty0 89--100, 1976.

\bibitem[Zadeh(1994)]{zadeh-1994-soft}
L.A. Zadeh.
\newblock Soft computing and fuzzy logic.
\newblock \emph{IEEE Software}, 11\penalty0 (6):\penalty0 48--56, 1994.
\newblock \doi{10.1109/52.329401}.

\end{thebibliography}
}

\newpage
\appendix
\section*{Appendix}

\section{Prompting Framework Details}
\label{appendix:framework}
\subsection{Symbolic Prompt Architecture}
Our framework integrates three symbolic control layers into the LLM prompting pipeline: (1) Boundary Prompts, (2) Fuzzy Scaffolding Schema, and (3) Symbolic Memory Schema. Together, these layers define a structured prompting architecture that supports context-aware, adaptive behavior without modifying the underlying model.

\textbf{1. Boundary Prompts}
The boundary prompt establishes the instructional scope and task-role framing. It defines the epistemic space within which the assistant operates (e.g., domain content, pedagogical tone, and role expectations). This natural language instruction serves as a high-level behavioral boundary for inference.

\textit{Example Snippet:}
\begin{lstlisting}
    You are an intelligent tutor that adapts your scaffolding strategies based on the student's grade level, task, and knowledge level.

---

### Student Info
- **Grade**: {grade}
- **Task**: {task}
- **Knowledge Level**: {label}
- **Strategies**:
  - {strategy_text}

---

### Scaffolding Logic (based on `fuzzy_logic_scaffolding.json`)
You must follow the rules and ranges from the `scaffolding_recipe.json` file to make decisions. Specifically:

1. Match the **task type** using `tasks.task_types` keywords.
2. Apply the **scaffolding strategy** from `scaffolding_settings`, based on the student's current knowledge level.
3. Select a **scaffolding_type** corresponding to the appropriate level of support.
4. Adjust your **vocabulary and complexity** using `readability_levels` based on the student's grade.
5. Monitor learning using `knowledge_update_rules`, including:
   - success/failure streaks
   - hint requests
   - time delays or confusion signals

---

### Short-Term Memory Instructions (`short_term_schema.json`)
You have access to a short-term memory file during the session. This memory stores important context such as:

- Learner misconceptions or strengths
- Concepts already explained
- Scaffolding types and strategies used so far
- Knowledge state progression
- Any misunderstandings, confusions, or repeated help requests

**When to update:**

- After identifying a new misconception, add it to `short_term_schema["misconceptions"]`.
- After a correct answer with confidence, record concept mastery in `short_term_schema["mastered_concepts"]`.
- After a repeated help request, escalate scaffolding level in `short_term_schema["scaffolding_history"]`.
- If the student expresses frustration or disengagement, flag in `short_term_schema["affective_state"]`.

**How to update:**

- Use structured JSON keys provided in the schema.
- Never overwrite past entries - always **append** new states or transitions.
- Reflect any key learner state changes in the memory schema before the next interaction.

---

### After Each Response
- Ask the student whether they understood.
- Offer help or suggest switching to another task.
- Adapt your behavior based on both the current scaffolding recipe **and** the contents of `short_term_schema.json`.

---

### Goal
Your job is not just to answer. You must **interpret**, **adapt**, and **track** the learner's state using both scaffolding logic and short-term memory.

\end{lstlisting}

\textbf{2. Fuzzy Scaffolding Schema}
The fuzzy scaffolding schema encodes heuristic rules for adapting instructional strategies under uncertainty. Unlike rigid logic trees, this schema represents learner states (e.g., confidence, confusion, engagement) as graded variables within fuzzy sets. It maps input cues to scaffolding strategies using interpretive ranges rather than binary thresholds.
Each fuzzy variable (e.g., \texttt{scaffolding\_settings}) is evaluated along a spectrum, and membership functions assign degrees of belief to each category. Rules in the \texttt{natural language boundary prompt}, allowing the model to make soft decisions even when input signals are ambiguous. These rules are interpretable and extensible, supporting adaptive strategy selection through simple, rule-based inference over learner signals parsed from the dialogue context.

\textit{Example Logic (simplified)}
\begin{lstlisting}
    {
  "tasks": {
    "task_types": [
      {
        "type": "recall",
        "description": "Retrieve basic facts or definitions.",
        "keywords": ["define", "identify", "list", "name", "label"]
      },
    ]
  },
  "scaffolding_settings": {
    "emerging": {
      "description": "Student is just beginning to understand the topic and needs high support.",
      "strategies": [
        "use step-by-step guidance",
        "provide worked examples",
        "ask yes/no or multiple choice questions",
        "reinforce understanding with visuals"
      ]    
  },
  "knowledge_levels": {
    "emerging": {
      "level": 1,
      "description": "User knows little or nothing about the topic. Identify this from phrases like: 'I'm not sure', 'I haven't learned this yet', 'Can you explain it from the start?'"
    },
  "scaffolding_types": {
    "high": {
      "description": "Aligns with 'emerging'. Provide extensive support and break down tasks."
    },
  "readability_levels": {
    "grade_1": {
      "flesch_kincaid_score": 90,
      "description": "Very easy to read. Suitable for early elementary students."
    },  
  "mappings": {
  "knowledge_levels": {
    "emerging": 1,
    "developing": 2,
    "proficient": 3,
    "advanced": 4
  },
  "scaffolding_types": {
    "high": 3,
    "moderate": 2,
    "low": 1,
    "challenge": 0
  },
  "readability_levels": {
    "grade_1": 90,
    "grade_3": 80,
    "grade_5": 70,
    "grade_8": 60,
    "grade_10": 50,
    "college": 30
  }
}
}

\end{lstlisting}

\textbf{3. Symbolic Memory Schema}
The symbolic memory schema maintains a lightweight, structured representation of the dialogue state across turns. It stores key session variables such as inferred learner confidence, prior strategies used, and current instructional goals. This schema is updated dynamically at the end of all user conversation turns, and can be queried in subsequent turns to support consistency and context-aware reasoning.

\textit{Example Entry:}
\begin{lstlisting}
    {
  "uid": "a15a85f6-536f-426f-b841-ec1f09daa1c1",
  "start_time": "2025-08-21T00:11:29.674983Z",
  "last_updated": "2025-08-22T16:24:04.564877",
  "task_type": "6",
  "knowledge_levels": 1,
  "scaffolding_type": 3,
  "readability_levels": 90,
  "grade": "11",
  "task_topic": "moon phases",
  "timestamp": "2025-08-21T00:35:54.009737",
  "user_provided_knowledge_level": true,
  "scaffolding_strategies": [
    "use open-ended questions",
    "prompt for reasoning and justification",
    "encourage connections to prior knowledge"
  ]
}
\end{lstlisting}

\section{Evaluation Rubric}
\label{appendix:rubric}
To assess cognitive adaptivity and instructional quality across prompting conditions, we implemented a rubric-based evaluation pipeline using GPT-4o. This rubric operationalizes core dimensions of tutoring competence inspired by educational psychology and cognitive modeling frameworks.

Each assistant response was evaluated along five key dimensions:
\begin{enumerate}
    \item Scaffolding quality: Quality of instructional support and progression.
    \item Contextual responsiveness: Sensitivity to learner input and conversational flow.
    \item Helpfulness: Clarity, accuracy, and pedagogical value of the explanation.
    \item Symbolic strategy use: Use of structured reasoning tools (e.g., analogies, decomposition).
    \item Memory of conversation: Coherence with earlier turns and accurate referencing.
\end{enumerate}

Each response was rated on a 1–5 scale per dimension, where 1 represents minimal competence and 5 indicates high proficiency. Ratings were accompanied by a free-text justification for transparency and interpretability.

Rather than relying on simple numerical scoring, we instructed GPT-4o to act as a cognitive science research assistant. The model received a structured rubric and sample anchor ratings for each criterion. For each assistant response, the model output a JSON object containing scores and a brief rationale.

Below is the scoring prompt used to guide the evaluation model:

\begin{lstlisting}
You are a research assistant evaluating dialogue responses from an AI system. You will rate each assistant response using the **RUBRIC** below, considering how well the assistant behaves like a helpful, cognitively competent collaborator.

**RUBRIC**
1. Scaffolding Quality (1-5):
    - Does the assistant guide the learner in a structured way, building progressively on their knowledge?
    - Is there evidence of follow-up, clarification, and adaptive instruction?
    - **Scoring Examples:**
        - 1 = No scaffolding or direction: Assistant dumps information or answers directly with no instructional structure.
        - 3 = Some guidance, but not adaptive: Offers isolated prompts or hints, but doesn't adjust based on learner input.
        - 5 = Clear, sophisticated, adaptive scaffolding that builds on learner's input: Guides the learner step-by-step, adjusts based on learner progress or confusion.

2. Contextual Responsiveness (1-5):
    - Does the assistant appropriately react to the learner's inputs and needs?
    - Does it acknowledge and respond to confusion or curiosity?
    - **Scoring Examples:**
        - 1 = Ignores learner input: Continues without reacting to previous statements, questions, or misunderstandings.
        - 3 = Acknowledges learner, but limited adaptation: Paraphrases or repeats but doesn't shift strategy in response to confusion or curiosity.
        - 5 = Fully responsive to learner's goals, confusion, or direction: Actively adjusts tone, content, or pacing based on learner's cues; tracks emotional/semantic signals.

3. Helpfulness (1-5):
    - Does the assistant clearly explain the concept or guide the learner toward understanding?
    - Is the information accurate and beneficial?
    - **Scoring Examples:**
        - 1 = Unhelpful or confusing: Responses are off-topic, incorrect, or misleading.
        - 3 = Somewhat useful explanation or prompt: Basic information or question posed, but lacks clarity or direction.
        - 5 = Clarifies the concept, supports learning effectively: Offers targeted, understandable, and pedagogically sound support.

4. Symbolic Strategy Use (1-5):
    - Does the assistant use structured reasoning strategies (e.g., analogies, scaffolds, labels, decomposition)?
    - **Scoring Examples:**
        - 1 = No structured reasoning (purely reactive): Just answers questions directly, no deeper structure or strategy.
        - 3 = Occasional heuristic or hint: Provides a simple strategy (e.g., "think about it in parts"), but without clear structure.
        - 5 = Consistent use of labels, analogy, abstraction, or scaffolding: Applies symbolic tools like metaphors, labeled steps, visual language, or structured breakdowns.

5. Memory of Past Conversation (1-5):
    - Does the assistant correctly reference or build upon earlier parts of the conversation?
    - Does it show awareness of prior learner questions, answers, or misunderstandings?
    - Are callbacks coherent and consistent with prior context?
    - **Scoring Examples:**
        - 1 = No memory: The assistant repeats prior explanations or contradicts itself, ignoring earlier learner input.
        - 3 = Partial memory: The assistant recalls some prior context (e.g., learner's question or a metaphor) but misses key misunderstandings or contradicts earlier turns.
        - 5 = Strong memory: The assistant effectively builds on earlier parts of the conversation, accurately references learner misconceptions or answers, and maintains coherent progression throughout.

Return your rating and justification in **valid JSON** format:
{
  "scaffolding_quality": 3,
  "contextual_responsiveness": 5,
  "helpfulness": 5,
  "symbolic_strategy_use": 3,
  "memory_conversation": 3,
  "justification": "The assistant responded appropriately, offered follow-up questions, and used a helpful analogy to illustrate the concept, though scaffolding could have been stronger."
}

Only return valid JSON.
\end{lstlisting}

\section{Ablation Mapping and Component Impact}
\label{appendix:ablation-map}

Table~\ref{tab:ablation-map} summarizes the contribution of each symbolic component by mapping ablation conditions to their functional roles and observed performance drops.

\begin{table}[h]
\centering
\caption{Ablation conditions and their observed effects}
\label{tab:ablation-map}
\begin{tabular}{p{2.5cm} p{3.2cm} p{3.2cm} p{5cm}}
\toprule
\textbf{Condition} & \textbf{Component Removed} & \textbf{Component Function} & \textbf{Observed Effect} \\
\midrule
C0 (Full) & None & All symbolic components present & Highest scores across all metrics \\
C1 (No-Memory) & Symbolic memory schema & Maintains dialogue context and misconceptions across turns & Major drop in \textit{Memory of Conversation}, slight drop in \textit{Scaffolding Quality} and \textit{Symbolic Strategy Use} \\
C2 (No-Fuzzy) & Fuzzy Heuristic Schema & Enables adaptive scaffolding under uncertainty via graded reasoning & Noticeable drop in \textit{Scaffolding Quality}, \textit{Contextual Responsiveness}, and \textit{Symbolic Strategy Use} \\
C3 (No-Boundary) & Boundary Prompt & Provides domain-specific role framing and tone regulation & Broad performance drop, especially in \textit{Helpfulness} and \textit{Contextual Responsiveness} \\
C4 (Vanilla) & All symbolic layers & Pure zero-shot prompting, no architectural support & Lowest scores across all metrics \\
\bottomrule
\end{tabular}
\end{table}

This mapping highlights how each component contributes uniquely to the model's ability to reason adaptively and maintain coherent tutoring behavior.

\section{Results Summary}
\label{appendix:results}
Table \ref{tab:mean-scores} reports the mean scores for each evaluation dimension across all five prompt conditions. Each score is averaged across 50 tutoring dialogues (25 per topic), with each assistant response rated using the 5-point rubric.

\begin{table}[h]
\centering
\caption{Mean Scores by Prompt Condition}
\label{tab:mean-scores}
\begin{tabular}{lccccc}
\toprule
\textbf{Condition} & \textbf{Scaffold} & \textbf{Responsive} & \textbf{Helpful} & \textbf{Symbolic} & \textbf{Memory} \\
\midrule
C0 (Full)         & 4.80 & 4.88 & 4.76 & 4.72 & 4.64 \\
C1 (No-Memory)    & 4.28 & 4.44 & 4.36 & 4.32 & 3.76 \\
C2 (No-Fuzzy)     & 4.24 & 4.40 & 4.28 & 4.00 & 3.92 \\
C3 (No-Boundary)  & 4.20 & 4.08 & 4.04 & 3.72 & 3.80 \\
C4 (Vanilla)      & 3.80 & 3.72 & 3.60 & 3.24 & 3.00 \\
\bottomrule
\end{tabular}
\end{table}

\subsection*{Chart Description: Average Evaluation Scores per Condition}

\begin{figure}{}
  \centering
  \includegraphics[width=0.9\linewidth]{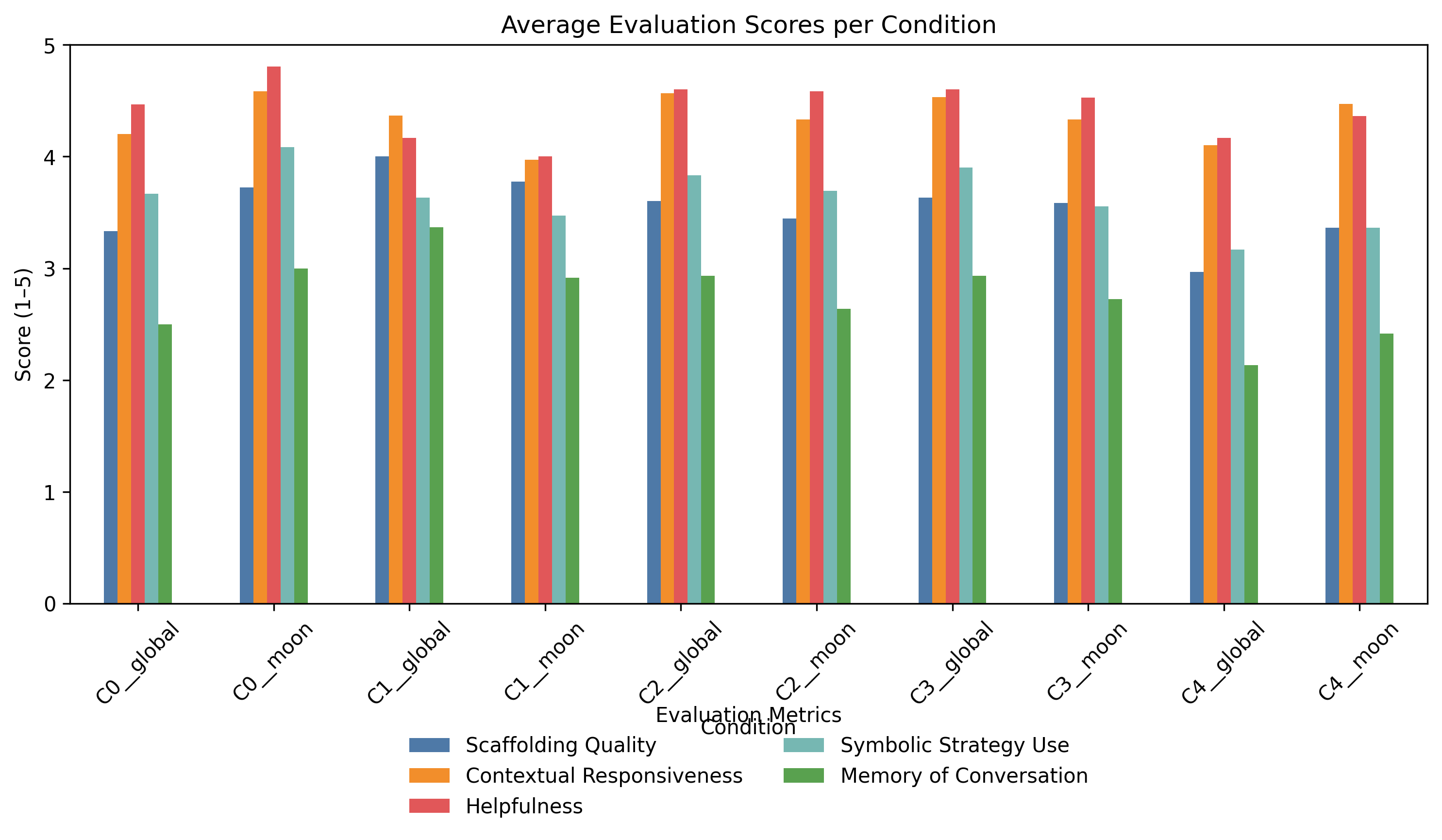}
  \caption{Bar chart of average evaluation scores across experimental conditions.}
  \label{fig:average_scores}
\end{figure}

Figure \ref{fig:average_scores} visualizes the \textbf{average evaluation scores} across \textbf{ten experimental conditions}. Each condition combines a specific \textit{model variant} (C0 to C4) and a \textit{topic domain} (``global'' or ``moon''). The x-axis denotes the experimental conditions, while the y-axis represents the average evaluation scores on a scale from \textbf{1 to 5}.

\paragraph{Evaluation Metrics (Bar Colors):}
Each colored bar corresponds to a distinct evaluation metric:
\begin{itemize}
    \item \textbf{Blue} – Scaffolding Quality
    \item \textbf{Orange} – Contextual Responsiveness
    \item \textbf{Red} – Helpfulness
    \item \textbf{Cyan} – Symbolic Strategy Use
    \item \textbf{Green} – Memory of Conversation
\end{itemize}

\paragraph{Condition Legend (X-Axis Labels):}
Each condition label follows the format \texttt{C\{n\}\_\{topic\}}, where:
\begin{itemize}
    \item \texttt{C0} – Full framework
    \item \texttt{C1} – No memory
    \item \texttt{C2} – No fuzzy logic
    \item \texttt{C3} – No boundary mechanism
    \item \texttt{C4} – Vanilla (baseline)
\end{itemize}

Each condition is evaluated under two topics:
\begin{itemize}
    \item \texttt{\_global} – Global Warming topic
    \item \texttt{\_moon} – Moon Phases topic
\end{itemize}

\paragraph{Observations:}
\begin{itemize}
    \item \textbf{C0 (Full framework)} yields the highest performance, particularly in \textit{Helpfulness} and \textit{Contextual Responsiveness}.
    \item \textbf{C4 (Vanilla)} exhibits the lowest scores, especially in \textit{Memory of Conversation}.
    \item Intermediate variants (C1 to C3) show a gradual performance drop as components are removed.
    \item \textit{Symbolic Strategy Use} and \textit{Memory of Conversation} metrics are notably impacted by ablations.
\end{itemize}

%%%%%%%%%%%%%%%%%%%%%%%%%%%%%%%%%%%%%%%%%%%%%%%%%%%%%%%%%%%%

\end{document}